\documentclass{bmvc2k}

\title{TAP-Out: Tracking Any Point \\ in 360\dg{} via Outpainting}

\addauthor{Finlay G. C. Hudson}{finlay.gc.hudson@york.ac.uk}{1}
\addauthor{James A. D. Gardner}{james.gardner@york.ac.uk}{1}
\addauthor{William A. P. Smith}{william.smith@york.ac.uk}{1}

\addinstitution{
 Department of Computer Science \\
University of York \\
York \\ 
United Kingdom
}

\runninghead{TAP-Out}{Tracking Any Point in 360\dg{} via Outpainting}

\usepackage{amsmath,amssymb,bm}
\usepackage{graphicx}
\usepackage{booktabs}
\usepackage{tikz}
\usetikzlibrary{calc}
\usepackage{caption}
\usepackage{multirow}
\usepackage[section]{placeins}
\usepackage[strings]{underscore}
\usepackage{cleveref}

\def\ie{\emph{i.e}\bmvaOneDot}

\newcommand{\dg}{\ensuremath{^\circ}}

\newcommand{\mypara}[1]{\noindent\textbf{#1}\hspace{0.1cm}}

\newcommand{\FH}[1]{}
\newcommand{\JG}[1]{}
\newcommand{\WS}[1]{}

\begin{document}

\maketitle

\begin{abstract}
Tracking points that leave the camera's field of view (FoV) requires reasoning about unobserved scene dynamics, a fundamentally different problem from tracking visible correspondences. We present TAP-Out, an offline method that makes this out-of-view (OOV) reasoning explicit by leveraging generative video outpainting as a world-model prior. Given a narrow-FoV input video, TAP-Out outpaints a complete 360\dg{} equirectangular video, locates target objects within the panoramic scene via video object segmentation, and synthesises object-centred perspective crops on which standard in-frame trackers operate. This decoupled design composes off-the-shelf foundation models without requiring task-specific architectural modifications. On the TAPVid360 benchmark, TAP-Out achieves state-of-the-art performance, improving OOV tracking accuracy by approximately $3\times$ over the best prior method. We demonstrate its flexibility on out-of-frame object segmentation and detection and we provide detailed diagnostic analyses across motion regimes, static versus dynamic content, and hallucination quality, characterising when and why the approach succeeds or fails. Code is publicly available at \url{https://github.com/finlay-hudson/TAP-Out}.
\end{abstract}

%-------------------------------------------------------------------------
\section{Introduction}
\label{sec:intro}

Humans and animals possess a remarkable ability to track dynamic objects even after those objects leave their field of view. This capability has been extensively studied in cognitive psychology under the framework of \emph{spatial updating}~\cite{klatzky2003encoding, giudice2011functional}. When moving through dynamic scenes, observers continuously update the egocentric locations of objects relative to their body, even when those objects are no longer visible~\cite{klatzky2003encoding, giudice2011functional}. Crucially, these maintained representations take the form of \emph{spatial images}, externalised, three-dimensional representations in working memory~\cite{loomis2013representing, loomis2012spatial} that persist after sensory input ceases and can represent locations in any direction around the observer~\cite{giudice2011functional}. We note that spatial updating is a \emph{motivating analogy} for our computational approach, not a claim of biological equivalence. The cognitive mechanisms underlying this ability remain debated: some evidence favours egocentric reference frames updated via proprioceptive feedback~\cite{klatzky1998allocentric, wang2016spatial}, while other work points to allocentric (world-centred) representations~\cite{burgess2006spatial, mou2004allocentric}, 3D scene graphs, or hybrid approaches. Our method draws conceptual inspiration from the idea of maintaining a complete spatial representation, regardless of the precise cognitive implementation.

\begin{figure}[t]
    \centering
    \begin{tikzpicture}
    \node (img) {\includegraphics[width=\textwidth]{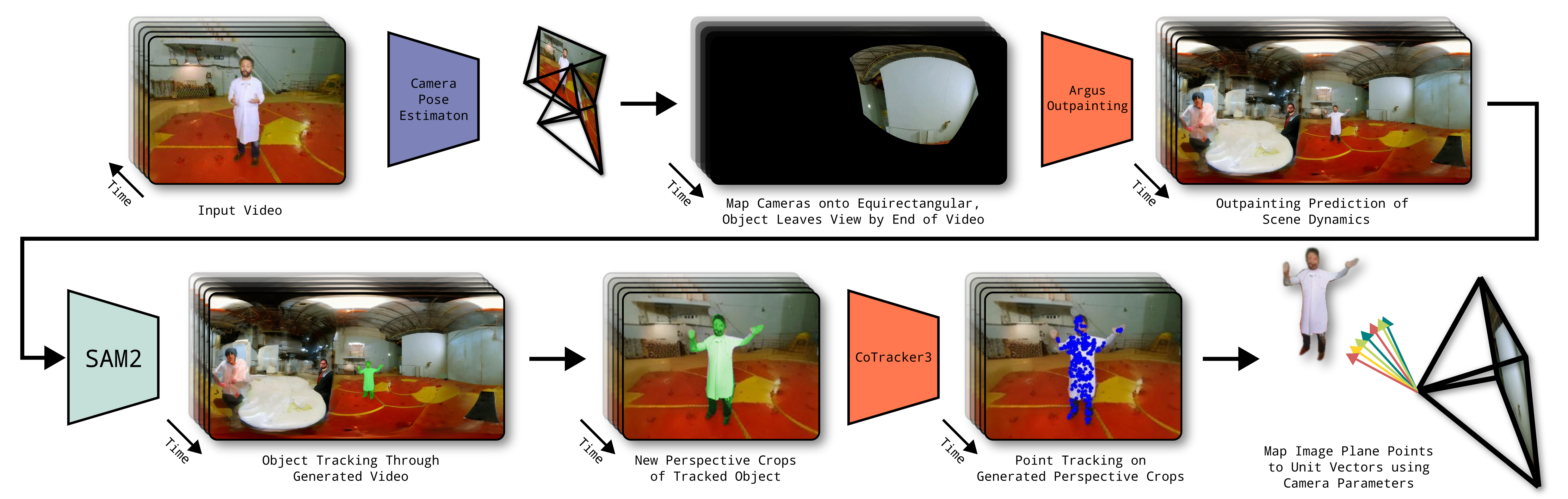}};
    \end{tikzpicture}
    \caption{TAP-Out pipeline overview. From a narrow field-of-view input video, we estimate camera poses and reproject frames onto an equirectangular canvas. A generative video outpainting model completes the full 360\dg{} scene, including unobserved regions. Given query points, we locate the target object in the panoramic video via video object segmentation, then synthesise object-centred perspective crops. Standard in-frame point tracking on these crops yields accurate 360\dg{} point tracks, even for points that have left the original field of view. This decoupled design enables any component to be independently upgraded as the field advances.}
    \label{fig:teaser}
\end{figure}

In computational vision, the Track Any Point (TAP) task densely tracks point correspondences over long ranges and through occlusions, with downstream applications in robotics~\cite{vecerik2024robotap, bharadhwaj2024track2act}, action recognition~\cite{kumar2024trajectory}, video editing~\cite{pan2023drag, wu2024draganything}, and medicine~\cite{azad2025taming, schmidt2024surgical}. Current 2D formulations~\cite{tapir2023, cotracker2024, tapnext2025} cannot, by definition, represent or track points that leave the camera frustum, yet this is the default behaviour for a tracker operating in a 3D world. A logical extension is 3D point tracks~\cite{tapip3d2025, tapvid3d2024}, but obtaining dense long-term 3D ground truth for dynamic OOV objects remains very challenging. Recently, the TAPVid-360 benchmark~\cite{hudson2025tapvid} was introduced, which requires tracking the egocentric 3D direction to a point \emph{even when it leaves the FoV}, leveraging 360\dg{} video data to create ground truth OOV tracks. The accompanying baseline, CoTracker360~\cite{hudson2025tapvid}, modifies CoTracker3~\cite{cotracker32024} to regress unit-sphere direction vectors. However, despite this specialisation, it performs poorly on OOV points: the model must simultaneously solve high-frequency local correspondence (for in-view points) and high-level world-model inference (for OOV points), capabilities that are very different and difficult to learn jointly from a tracking loss alone.

In this work, we advocate for decoupling these two problems. We propose TAP-Out, a modular pipeline that uses a standard off-the-shelf 2D tracker across two operational states. When objects are in view, the tracker consumes high-fidelity original frames. When objects move out of view, we reframe tracking as a scene completion problem: a powerful pre-trained generative video model acts as a world-model prior, outpainting the full 360\dg{} scene so that the target is always visible to the tracker. This design offers three practical advantages. First, it requires no custom 360-specific architectures, relying instead on modular composition of existing models. Second, it inherits progress from generative modelling: as video generation improves, our OOV tracking improves without retraining. Third, despite this simplicity, it substantially outperforms the specialised CoTracker360.

A key insight is that \emph{all} OOV tracking methods must extrapolate beyond observed data, a form of implicit hallucination. CoTracker360, for instance, must predict plausible directions for points it has never observed, but does so without any explicit visual justification. TAP-Out makes this extrapolation explicit: the generative model produces a complete visual scene consistent with the observed video, providing a human-interpretable explanation of \emph{why} each point is believed to lie in a particular direction. Even if the generated content is imperfect, this explicit visual hallucination outperforms implicit direction prediction, as our results demonstrate.

We note that TAP-Out is an \emph{offline} method, processing the complete video before producing tracks. This is suited to applications where track quality supersedes speed, such as visual effects (VFX), video editing, forensic video analysis, and sports analytics. Runtime is approximately 470 seconds per video compared to 0.15 seconds for CoTracker360; we view speed improvements via distillation as promising future work.

Our contributions are as follows:
\begin{itemize}
\item A new paradigm for 360\dg{} point tracking that formally decouples in-view correspondence from OOV generation.
\item \textbf{TAP-Out}, a modular system that composes off-the-shelf foundation models with a pre-trained generative video world-model prior, providing the first empirical evidence that generative video models can serve as effective priors for OOV tracking.
\item State-of-the-art results on TAPVid-360, achieving an approximate threefold improvement on dynamic OOV points compared to existing methods, alongside comprehensive diagnostic analyses across motion regimes, static versus dynamic content, and hallucination quality.
\end{itemize}

%-------------------------------------------------------------------------
\section{Related Work}
\label{sec:related_work}

\mypara{Amodal Completion.}
Amodal completion estimates the full shape or appearance of objects including unseen parts. Image-based methods use spatial context from a single frame to complete occluded regions~\cite{ozguroglu2024pix2gestalt, zhan2024amodal}, while temporal methods leverage cues across sequential frames~\cite{vanhoorick2023tcow, hudson2024track, chen2025using}. The most challenging form is out-of-view completion, where an object has entirely left the sensor's FoV, directly related to human spatial updating~\cite{klier2008spatial, dordevic2022role}. CoTracker360~\cite{hudson2025tapvid} addresses this with promising but limited results, as joint learning of camera motion and spatial understanding constrains generalisation.

\mypara{Video Generation and World Models.}
Large-scale generative video models such as Sora~\cite{openai2024video} and VEO~\cite{GoogleDeepMind_Veo} have demonstrated remarkable zero-shot capabilities~\cite{wiedemer2025video}, suggesting the acquisition of implicit world models that capture object dynamics and physical plausibility. These capabilities are being applied to tasks including amodal completion~\cite{hudson2024track, chen2025using}, robotic manipulation~\cite{patel2025robotic}, self-supervised learning~\cite{belagali2024gen}, and autonomous driving~\cite{bartoccioni2025vavim}. The Argus model~\cite{beyondtheframe360outpainting2025} is particularly relevant: it trains a video diffusion model to produce temporally consistent equirectangular video from perspective input, explicitly estimating scene dynamics beyond the camera's FoV. Our work provides the first empirical demonstration that such generative models can serve as effective world-model priors for OOV point tracking, a capability not previously explored in the 2D, 3D, or 360\dg{} tracking literature.

\mypara{Camera Estimation.}
Accurate camera parameters are essential for mapping perspective frames onto the spherical plane. Recent feed-forward methods such as VGGT~\cite{wang2025vggt}, MASt3R~\cite{leroy2024grounding}, and MegaSaM~\cite{li2025megasam} directly regress 3D scene attributes with strong performance. We evaluate our pipeline with both ground-truth and estimated cameras to characterise the impact of estimation errors on downstream tracking.

\mypara{Track Any Point and 3D Tracking.}
The TAP task~\cite{harley2022particle} extends classical optical flow~\cite{baker2004lucas, Barron1994, dosovitskiy2015flownet, teed2020raft} to long-term, occlusion-robust tracking, with benchmarks~\cite{tapvid2022} and models including TAPIR~\cite{tapir2023}, CoTracker~\cite{cotracker2024, cotracker32024}, and TAPNext~\cite{tapnext2025}. These 2D methods cannot represent OOV points by construction.

3D tracking methods~\cite{xiao2024spatialtracker, wang2024scenetracker, tapvid3d2024, tapip3d2025} estimate trajectories in 3D space but typically focus on points within the visible frustum. Extending them to OOV is fundamentally ill-posed: estimating the 3D position (including depth) of an unobserved point requires strong scene priors that current 3D trackers lack. The depth component is especially ambiguous outside the frustum, where no visual evidence constrains the estimate. In contrast, the 360\dg{} formulation~\cite{hudson2025tapvid} requires only direction (on $S^2$), sidestepping the depth ambiguity. This is well-matched to generative video models that operate in pixel space and can leverage massive 2D video pretraining to answer ``where is the object?'' (direction) without solving ``how far away is it?'' (depth). The TAPVid-360-10k benchmark~\cite{hudson2025tapvid} provides large-scale evaluation data by sampling perspective camera trajectories from 360\dg{} video and modifying CoTracker3~\cite{cotracker32024} to regress unit-sphere direction vectors as the CoTracker360 baseline.

%-------------------------------------------------------------------------
\section{Method}
\label{sec:method}

Given an input perspective video $\mathcal{V}_{\text{pers}} = \{I_t\}_{t=1}^T$ consisting of $T$ RGB frames $I_t \in \mathbb{R}^{3 \times H \times W}$ and a set of query points $\mathcal{Q} = \{q_k\}_{k=1}^K$ specified in the first frame, our goal is to track these points throughout the video, even when they move outside the camera's FoV. Unlike traditional TAP methods that predict 2D pixel coordinates $p_t^k = (u_t^k, v_t^k)$, we adopt the TAPVid-360 formulation~\cite{hudson2025tapvid} and predict 3D direction vectors $\mathbf{d}_t^k \in \mathbb{R}^3$ with $\|\mathbf{d}_t^k\| = 1$ in the camera coordinate system for each query point at each frame.

Our method transforms this challenging OOV tracking problem into a standard in-frame tracking task through a series of geometric transformations and generative modelling. \Cref{fig:teaser} provides an overview of the pipeline.

\subsection{Coordinate Systems and Transformations}

We work with two primary coordinate representations: perspective images and equirectangular panoramas. We define a shared world coordinate frame aligned with the first camera frame, \ie $\mathbf{R}_1 = \mathbf{I}$, where $\mathbf{I}$ is the identity matrix. Placing all frames in a single consistent coordinate frame yields a consistent scene arrangement that simplifies the outpainting model's task. Each frame $I_t$ is associated with a camera rotation $\mathbf{R}_t \in SO(3)$ relative to the first frame and intrinsic parameters $\mathbf{K}_t$. Since we reason only about direction and not 3D position, translation vectors are discarded.

\mypara{Perspective to Equirectangular Projection.}
To map a pixel coordinate $(u, v)$ from the perspective image to the equirectangular representation, we first unproject it to a 3D ray in camera coordinates via normalised device coordinates:
\begin{equation}
\mathbf{p}_{\text{ndc}} = \begin{bmatrix}
\frac{2u}{W} - 1 \\
\frac{2v}{H} - 1 \\
1
\end{bmatrix}.
\end{equation}

Given horizontal and vertical FoVs $\alpha$ and $\beta$, the 3D direction vector in camera coordinates is:
\begin{equation}
\mathbf{d}_{\text{cam}} = \begin{bmatrix}
\mathbf{p}_{\text{ndc}}[0] \cdot \tan(\alpha/2) \\
\mathbf{p}_{\text{ndc}}[1] \cdot \tan(\beta/2) \\
-1
\end{bmatrix}.
\end{equation}

Transforming to the shared world frame using the inverse camera rotation:
\begin{equation}
\mathbf{d}_{\text{world}} = \mathbf{R}_t^{\top} \mathbf{d}_{\text{cam}}.
\end{equation}

We then convert to spherical coordinates $(\theta, \phi)$:
\begin{equation}
\theta = \text{atan2}({d}_{\text{world},y}, {d}_{\text{world},x}), \quad
\phi = \arcsin\left(\frac{{d}_{\text{world},z}}{\|\mathbf{d}_{\text{world}}\|}\right),
\end{equation}
and map to equirectangular pixel coordinates $(u_{\text{eq}}, v_{\text{eq}})$ in an image of dimensions $H_{\text{eq}} \times W_{\text{eq}}$:
\begin{equation}
u_{\text{eq}} = \frac{W_{\text{eq}}}{2\pi}(\theta + \pi), \quad
v_{\text{eq}} = \frac{H_{\text{eq}}}{\pi}\left(\frac{\pi}{2} - \phi\right).
\end{equation}

\subsection{TAP-Out Pipeline}

\mypara{Input Preprocessing.}
Given the input video $\mathcal{V}_{\text{pers}}$, we first estimate or obtain the camera parameters $\{\mathbf{R}_t, \mathbf{K}\}_{t=1}^T$ relative to the first frame, where $\mathbf{R}_1 = \mathbf{I}$. When ground-truth parameters are unavailable, we employ VGGT~\cite{wang2025vggt}. Using the projection described above, we map each perspective frame onto a shared equirectangular canvas, producing a masked equirectangular video $\mathcal{V}_{\text{eq}}^{\text{input}} = \{I_t^{\text{eq}}\}_{t=1}^T$ where $I_t^{\text{eq}} \in \mathbb{R}^{3 \times H_{\text{eq}} \times W_{\text{eq}}}$. Regions outside the original FoV are masked. The initial query points $\mathcal{Q}$ are similarly transformed:
\begin{equation}
q_k^{\text{eq}} = \Pi_{\text{eq}}(q_k; \mathbf{I}, \mathbf{K}),
\end{equation}
where $\Pi_{\text{eq}}(\cdot)$ denotes the perspective-to-equirectangular projection operator.

\mypara{360\dg{} Scene Completion.}
We leverage Argus~\cite{beyondtheframe360outpainting2025}, a diffusion-based video-to-360 model, to complete the masked equirectangular video:
\begin{equation}
\mathcal{V}_{\text{eq}}^{\text{full}} = \text{Argus}(\mathcal{V}_{\text{eq}}^{\text{input}}).
\end{equation}
This model predicts both spatial layout and object dynamics in unobserved regions, providing a temporally consistent panoramic representation. To preserve high-frequency input details, we create a composite video:
\begin{equation}
\tilde{I}_t^{\text{eq}} = M_t \odot I_t^{\text{eq}} + (1 - M_t) \odot \hat{I}_t^{\text{eq}},
\end{equation}
where $\hat{I}_t^{\text{eq}}$ denotes a frame from $\mathcal{V}_{\text{eq}}^{\text{full}}$, $M_t$ is the binary mask indicating observed regions, and $\odot$ denotes element-wise multiplication.

\mypara{Object Localisation in Equirectangular Space.}
Given the complete equirectangular video and the initial query $q_k^{\text{eq}}$, we employ SAM2~\cite{ravi2024sam} to track the query object:
\begin{equation}
\mathcal{S} = \{S_t\}_{t=1}^T = \text{SAM2}(\{\tilde{I}_t^{\text{eq}}\}_{t=1}^T, q_1^{\text{eq}}),
\end{equation}
where $S_t \in \{0,1\}^{H_{\text{eq}} \times W_{\text{eq}}}$ is a binary segmentation mask. Although equirectangular videos are out-of-domain for SAM2, it generalises sufficiently to provide coarse masks for approximate object localisation. From each mask, we compute the object centroid:
\begin{equation}
c_t^{\text{eq}} = \frac{1}{|S_t|}\sum_{(u,v) \in S_t} \begin{bmatrix} u \\ v \end{bmatrix}.
\end{equation}

\mypara{Object-Centred Perspective Generation.}
For each frame, we generate a new perspective view centred on the tracked object. We convert the centroid $c_t^{\text{eq}}$ to a 3D direction:
\begin{equation}
\mathbf{d}_t^{\text{obj}} = \Pi_{\text{eq}}^{-1}(c_t^{\text{eq}}),
\end{equation}
and define a virtual camera with rotation $\tilde{\mathbf{R}}_t$ aligning its optical axis with $\mathbf{d}_t^{\text{obj}}$ and specified FoV $(\tilde{\alpha}, \tilde{\beta})$:
\begin{equation}
\tilde{I}_t^{\text{pers}} = \Pi_{\text{pers}}(\tilde{I}_t^{\text{eq}}; \tilde{\mathbf{R}}_t, \tilde{\alpha}, \tilde{\beta}).
\end{equation}
This produces object-centred perspective frames $\{\tilde{I}_t^{\text{pers}}\}_{t=1}^T$ where the query object is consistently visible and centred.

\mypara{360\dg{} Point Tracking.}
We apply a standard tracker such as CoTracker3~\cite{cotracker32024} to obtain refined tracks:
\begin{equation}
\{\tilde{p}_t^k\}_{t=1}^T = \text{Tracker}(\{\tilde{I}_t^{\text{pers}}\}_{t=1}^T, \tilde{q}^k),
\end{equation}
where $\tilde{q}^k$ is the query point in centred-view coordinates. The tracked points are then transformed back to the original camera coordinate system: (1)~convert to equirectangular coordinates $p_t^{k,\text{eq}} = \Pi_{\text{eq}}(\tilde{p}_t^k; \tilde{\mathbf{R}}_t, \tilde{\mathbf{K}})$, (2)~convert to a 3D direction in the shared world frame $\mathbf{d}_t^{k,\text{world}} = \Pi_{\text{eq}}^{-1}(p_t^{k,\text{eq}})$, and (3)~transform to the original camera frame $\mathbf{d}_t^k = \mathbf{R}_t \mathbf{d}_t^{k,\text{world}}$.

The pipeline is modular by design: any component (camera estimation, generative outpainting, segmentation, or point tracker) can be independently upgraded as the field advances. This ``future-proofing'' means that improvements in video generation directly translate to better OOV tracking without requiring retraining or architectural changes.

\subsection{Extension to Other Tasks}

Once we have the object-centred video $\{\tilde{I}_t^{\text{pers}}\}_{t=1}^T$ and corresponding query points, we can solve several related tracking tasks. For video object segmentation, the coarse equirectangular masks can be refined by re-running SAM2 on the object-centred narrow-FoV frames. For object detection, bounding boxes can be computed in the centred view and transformed back. In both cases, outputs are reprojected to the original camera coordinates via:
\begin{equation}
\hat{\mathbf{c}} = \pi(\mathbf{K} \cdot \mathbf{R}_t \cdot \Pi_{\text{eq}}^{-1}(\Pi_{\text{eq}}(\mathbf{c}; \tilde{\mathbf{R}}_t, \tilde{\mathbf{K}}))),
\end{equation}
where $\pi(\mathbf{x}) = (\mathbf{x}_1/\mathbf{x}_3, \mathbf{x}_2/\mathbf{x}_3)$ denotes perspective division.

%-------------------------------------------------------------------------
\section{Evaluation}
\label{sec:eval}

\subsection{Experimental Setup}

\mypara{Data.}
We evaluate TAP-Out on the TAPVid360-10k benchmark~\cite{hudson2025tapvid}, which provides perspective frames with ground-truth OOV information including world-to-camera rotations and focal lengths. Since Argus~\cite{beyondtheframe360outpainting2025} was trained on data partially overlapping with TAPVid360-10k, we isolated a non-overlapping subset of 8{,}887 videos, termed \textit{TAPVid360-TAPOut}.

\mypara{Point Categorisation.}
To isolate the contribution of the generative world model from simple geometric reprojection, we categorise points based on their intrinsic motion. Since the TAPVid360-10k benchmark provides ground-truth camera rotations, tracking static points is trivial: such points can be precisely located by reprojecting their first-frame positions into subsequent views using only the known camera motion. To move beyond this baseline and evaluate the generative model's capacity to predict actual object dynamics, we define \textbf{dynamic points} as those with cumulative world-frame motion exceeding $10^\circ$ over the video sequence. This threshold ensures that points exhibit significant intrinsic motion beyond typical camera estimation jitter or sensor noise, specifically capturing objects moving relative to the background. Points below this threshold are considered \textbf{static} or near-static.

\mypara{Metrics.}
We adopt the TAPVid360-10k metrics~\cite{hudson2025tapvid}. \textbf{$< \delta^{x}_{\mathrm{avg}}$}: the fraction of predicted points within a threshold angular distance of the ground truth (higher is better). \textbf{$AD^{x}_{\mathrm{avg}}$}: the mean angular distance in degrees between predicted and ground-truth points (lower is better). Both metrics are computed separately for points that are in-frame (\textbf{if}), out-of-frame (\textbf{oof}), and for all points combined (\textbf{all}).

\mypara{Implementation Details.}
TAP-Out uses four off-the-shelf foundation models: VGGT~\cite{wang2025vggt} for camera estimation (when ground truth is unavailable), Argus~\cite{beyondtheframe360outpainting2025} for 360\dg{} outpainting, SAM2~\cite{ravi2024sam} for object segmentation, and CoTracker3~\cite{cotracker32024} (offline mode) for point tracking. All experiments use a single NVIDIA A100 GPU. Total inference time is approximately \textbf{470 seconds per video} for TAP-Out, compared to \textbf{0.15 seconds} for CoTracker360. This ${\sim}3{,}000\times$ latency difference reflects our design choice: TAP-Out is an offline, high-precision tool for applications such as VFX, video editing, and forensic analysis, where track quality takes priority over speed.

\subsection{Main Results}

\begin{table}[!t]
\centering
\resizebox{\textwidth}{!}{
\begin{tabular}{lcccccc}
\toprule
Method & $< \delta^{x}_{\mathrm{avg}}$\textsubscript{all} $\uparrow$ & $< \delta^{x}_{\mathrm{avg}}$\textsubscript{if} $\uparrow$ & $< \delta^{x}_{\mathrm{avg}}$\textsubscript{oof} $\uparrow$ & ${AD^{x}_\mathrm{avg}}$\textsubscript{all}$\downarrow$ & ${AD^{x}_\mathrm{avg}}$\textsubscript{if} $\downarrow$ & ${AD^{x}_\mathrm{avg}}$\textsubscript{oof} $\downarrow$\\
\midrule
TAPNext \cite{tapnext2025} & 0.0075 {\tiny $\pm$0.0077} & 0.0178 {\tiny $\pm$0.0180} & 0.0004 {\tiny $\pm$0.0010} & 52.6330 {\tiny $\pm$18.4256} & 37.2676 {\tiny $\pm$19.6733} & 62.6947 {\tiny $\pm$19.8463} \\
TAPIR \cite{tapir2023} & 0.0096 {\tiny $\pm$0.0096} & 0.0229 {\tiny $\pm$0.0225} & 0.0003 {\tiny $\pm$0.0006} & 50.7070 {\tiny $\pm$18.4413} & 35.0867 {\tiny $\pm$21.8100} & 60.9705 {\tiny $\pm$17.8270} \\
BootsTAPIR \cite{bootstap2024} & 0.0106 {\tiny $\pm$0.0106} & 0.0254 {\tiny $\pm$0.251} & 0.0005 {\tiny $\pm$0.0011} & 49.1382 {\tiny $\pm$18.5384} & 34.4510 {\tiny $\pm$21.4859} & 58.6180 {\tiny $\pm$18.2366} \\
TAPIP3D \cite{tapip3d2025} & 0.2514 {\tiny $\pm$0.1606} & 0.4773 {\tiny $\pm$0.2429} & 0.0897 {\tiny $\pm$0.1301} & 35.9088 {\tiny $\pm$26.7414} & 22.9776 {\tiny $\pm$23.0838} & 44.8629 {\tiny $\pm$32.1248} \\
CoTracker3 (offline) \cite{cotracker2024} & 0.2375 {\tiny $\pm$0.0899} & 0.5516 {\tiny $\pm$0.1588} & 0.0146 {\tiny $\pm$0.0208} & 38.2831 {\tiny $\pm$21.9041} & 18.4400 {\tiny $\pm$22.2117} & 51.6813 {\tiny $\pm$24.4290} \\
CoTracker360 \cite{hudson2025tapvid} & 0.2259 {\tiny $\pm$0.1059} & 0.3925 {\tiny $\pm$0.1326} & 0.1062 {\tiny $\pm$0.0984} & 8.6054 {\tiny $\pm$7.6706} & 4.1213 {\tiny $\pm$4.8057} & 11.3619 {\tiny $\pm$9.8862} \\
Static Reprojection (Gt Camera) & 0.2497 {\tiny $\pm$0.1989} & 0.3192 {\tiny $\pm$0.2177} & 0.2032 {\tiny $\pm$0.2030} & 8.5689 {\tiny $\pm$10.9995} & 5.8791 {\tiny $\pm$7.4396} & 10.2048 {\tiny $\pm$13.5424} \\
\midrule
TAP-Out (GT Camera) & \textbf{0.4418} {\tiny $\pm$0.1769} & \textbf{0.6038} {\tiny $\pm$0.2103} & \textbf{0.3318} {\tiny $\pm$0.1794} & \textbf{4.0788} {\tiny $\pm$7.2187} & \textbf{2.5947} {\tiny $\pm$6.8070} & \textbf{5.0180} {\tiny $\pm$7.9054} \\
TAP-Out (VGGT Est Camera) & 0.2960 {\tiny $\pm$0.1499} & 0.5296 {\tiny $\pm$0.2126} & 0.1354 {\tiny $\pm$0.1314} & 17.4218 {\tiny $\pm$17.3905} & 8.0179 {\tiny $\pm$11.9628} & 23.2742 {\tiny $\pm$23.5585} \\
\bottomrule
\end{tabular}
}
\vspace{3mm}
\caption{Tracking performance on TAPVid360-TAPOut. Subscripts denote in-frame (\textbf{if}), out-of-frame (\textbf{oof}), and all points (\textbf{all}).}
\label{tab:tracking_performance_comparison}
\end{table}

\Cref{tab:tracking_performance_comparison} presents the main results. TAP-Out with ground-truth (GT) cameras achieves state-of-the-art performance across all metrics. The most pronounced advantage is on OOV points: $\delta^{x}_{\mathrm{avg,oof}} = 0.332$, approximately $3.1\times$ higher than CoTracker360's 0.106. This validates that explicit scene completion produces more accurate OOV tracks than implicit direction prediction.

While these results are strong, they represent the performance across the entire benchmark. As established in \Cref{sec:eval}, tracking static points is trivial given GT camera rotations. Therefore, the more rigorous test of our generative world model lies in its performance on dynamic objects, where geometric reprojection provides no solution.

We also report results with estimated cameras (VGGT). With estimated cameras, TAP-Out achieves $\delta^{x}_{\mathrm{avg,oof}} = 0.135$, exceeding CoTracker360 while substantially outperforming it on in-frame and overall metrics. The higher $AD^{x}_{\mathrm{avg}}$ indicates that while camera estimation is frequently accurate, failures can be catastrophic, skewing mean error. This highlights camera estimation as a key bottleneck for deployment.

\begin{table}[!t]
\centering
\begin{tabular}{lccc}
\toprule
\textbf{Regime} & Low ($0$\dg--$5$\dg) & Medium ($5$\dg--$30$\dg) & High ($30$\dg+) \\
\midrule
\% Points in Dataset & 0.8\% & 59.6\% & 39.6\% \\
TAP-Out $< \delta^{x}_{\mathrm{avg}}$\textsubscript{oof} & 0.90 & 0.45 & 0.21 \\
\bottomrule
\end{tabular}
\vspace{3mm}
\caption{TAP-Out OOV tracking accuracy ($< \delta^{x}_{\mathrm{avg,oof}}$) across camera motion magnitudes, measured as total rotation over the video clip.}
\label{tab:motion_regimes}
\end{table}

\begin{figure}[!t]
    \centering
    \resizebox{\textwidth}{!}{%
    \begin{tikzpicture}
        % LEFT COLUMN: Rows 1-4 (t=0 to t=15)
        \node[anchor=north west, inner sep=0] (left_img) at (0,0) {
            \includegraphics[width=5cm, clip, trim=30bp 785bp 0bp 23bp]{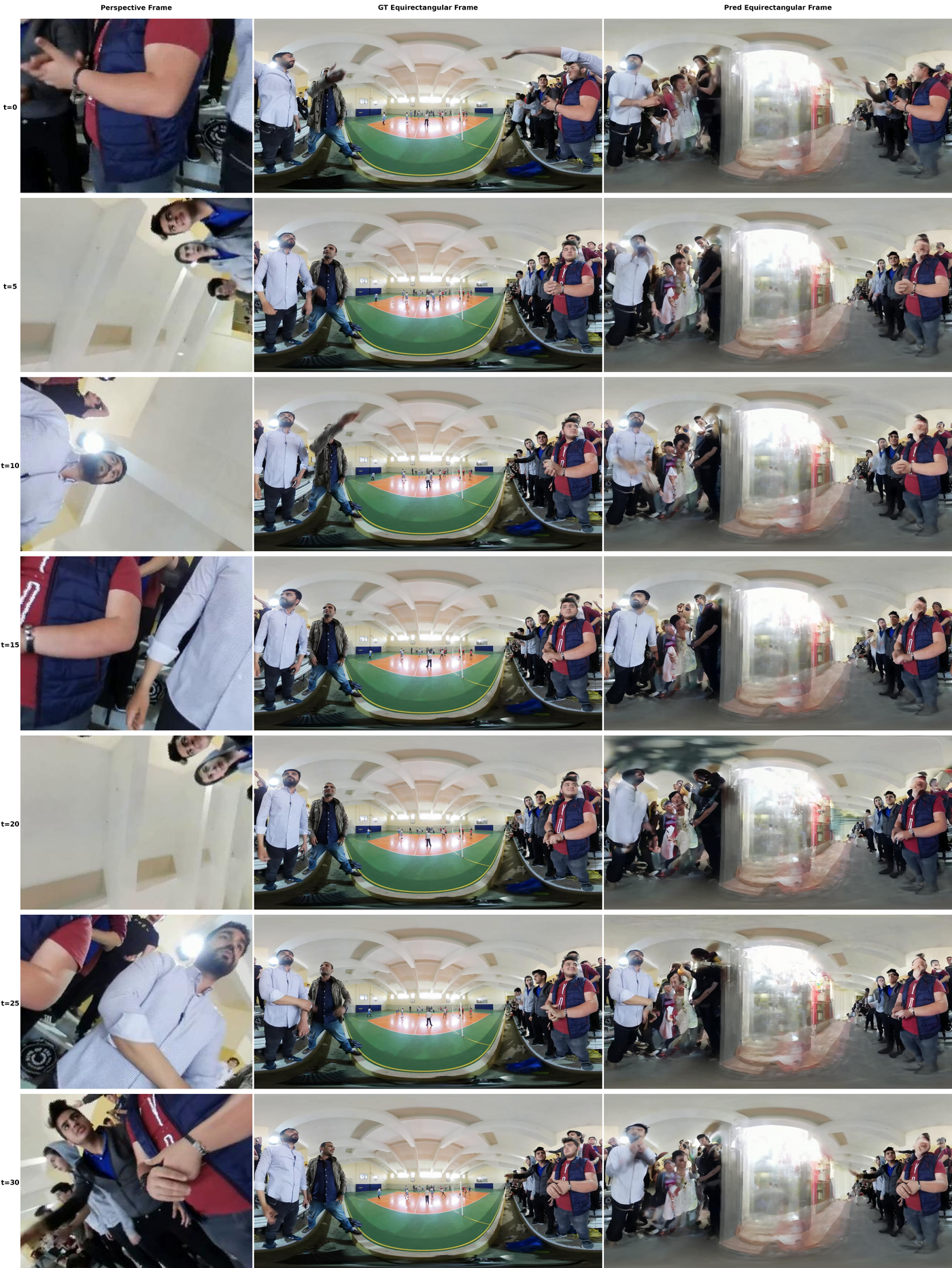}
        };
        % LEFT HEADERS
        \node[above, font=\bfseries\tiny] at ($(left_img.north) + (-1.85, -0.1)$) {Perspective};
        \node[above, font=\bfseries\tiny] at ($(left_img.north) + (-0.35, -0.1)$) {GT Equi};
        \node[above, font=\bfseries\tiny] at ($(left_img.north) + (1.55, -0.1)$) {Pred Equi};
        % LEFT ROW LABELS
        \node[left, font=\tiny, inner sep=1pt] at ($(left_img.north west)!0.125!(left_img.south west)$) {\textbf{t=0}};
        \node[left, font=\tiny, inner sep=1pt] at ($(left_img.north west)!0.375!(left_img.south west)$) {\textbf{t=5}};
        \node[left, font=\tiny, inner sep=1pt] at ($(left_img.north west)!0.625!(left_img.south west)$) {\textbf{t=10}};
        \node[left, font=\tiny, inner sep=1pt] at ($(left_img.north west)!0.875!(left_img.south west)$) {\textbf{t=15}};
        % RIGHT COLUMN: Rows 5-7 (t=20 to t=30)
        \node[anchor=north west, inner sep=0] (right_img) at ($(left_img.north east) + (0.5, 0)$) {
             \includegraphics[width=5cm, clip, trim=30bp 0bp 0bp 1075bp]{figures/combined_eq_for_rebuttal.png}
        };
        % RIGHT HEADERS
        \node[above, font=\bfseries\tiny] at ($(right_img.north) + (-1.85, -0.1)$) {Perspective};
        \node[above, font=\bfseries\tiny] at ($(right_img.north) + (-0.35, -0.1)$) {GT Equi};
        \node[above, font=\bfseries\tiny] at ($(right_img.north) + (1.55, -0.1)$) {Pred Equi};
        % RIGHT ROW LABELS
        \node[left, font=\tiny, inner sep=1pt] at ($(right_img.north west)!0.166!(right_img.south west)$) {\textbf{t=20}};
        \node[left, font=\tiny, inner sep=1pt] at ($(right_img.north west)!0.5!(right_img.south west)$) {\textbf{t=25}};
        \node[left, font=\tiny, inner sep=1pt] at ($(right_img.north west)!0.833!(right_img.south west)$) {\textbf{t=30}};
    \end{tikzpicture}
    }
    \vspace{3mm}
    \caption{Hallucination quality over time. Each row shows the original perspective view, the ground-truth equirectangular panorama, and the predicted (outpainted) equirectangular panorama at different timesteps.}
    \label{fig:hallucination_robustness}
\end{figure}

\subsection{Static Reprojection Baseline}
\label{sec:static_baseline}

A natural question is whether generative outpainting is necessary at all: if we have camera poses, can we simply reproject points geometrically? To answer this, we implemented a static reprojection baseline that takes the first visible position of each query point, assumes the point is static in the world, and reprojects it into subsequent frames using ground truth camera rotations.

As shown in \Cref{tab:tracking_performance_comparison}, the static baseline achieves an out-of-view accuracy of $\delta^{x}_{\mathrm{avg,oof}} = 0.203$, which is a significant improvement over 2D trackers but still lags far behind TAP-Out's $0.332$. While reprojection is effective for stationary background points, it is fundamentally unable to account for the dynamic object motion that characterizes the majority of the TAPVid360 benchmark. This confirms that while camera pose estimation provides a necessary geometric foundation, a generative world model is essential for reasoning about object dynamics and permanence during out-of-view tracking.

\begin{figure}[!t]
    \centering
    \begin{tikzpicture}
    \node (img) {\includegraphics[width=0.9\textwidth, clip, trim=0 0 0 1.5cm]{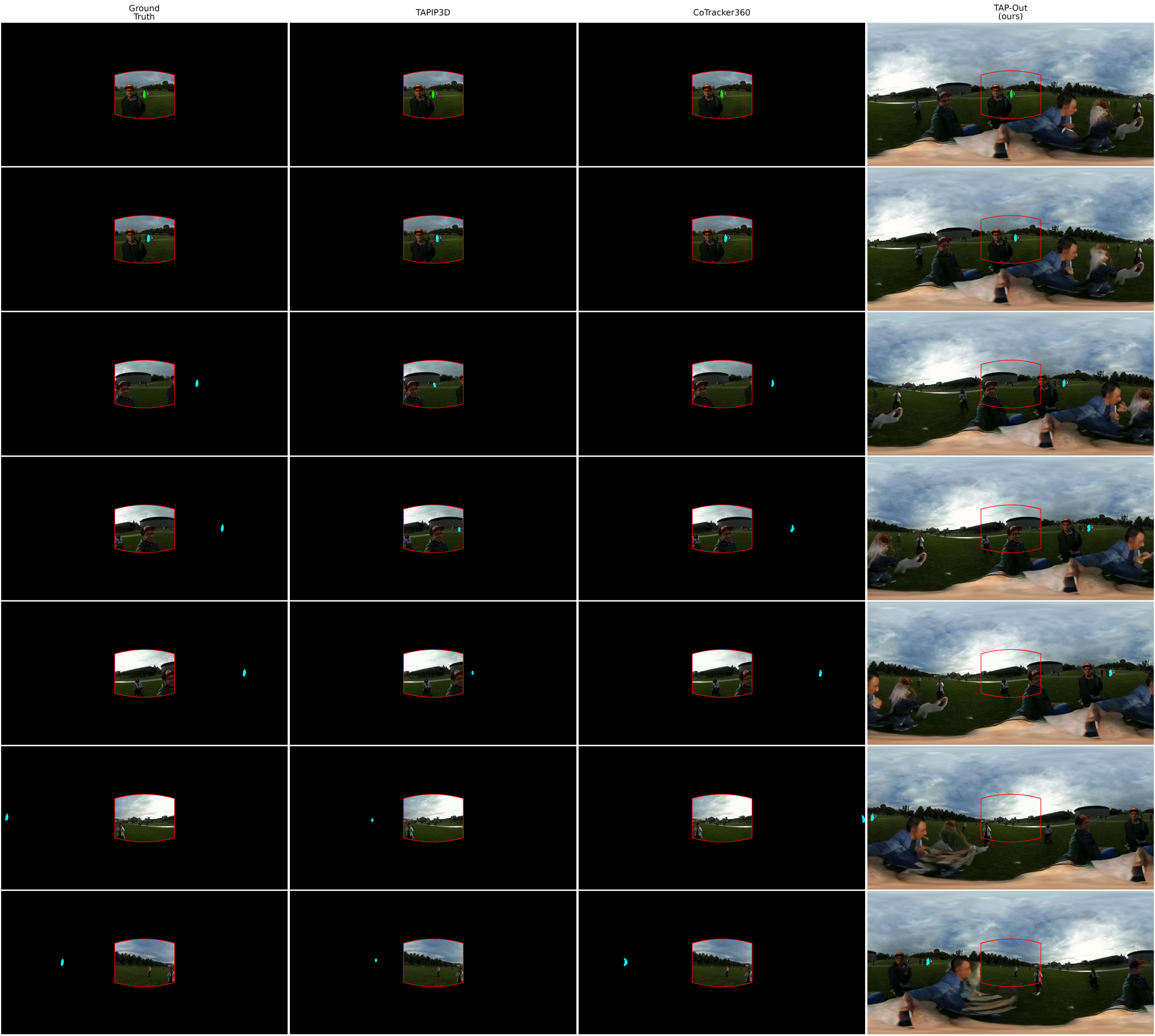}};
    \node at (-4.43, 5.25) [rotate=-0, align=center, font=\scriptsize] {Ground Truth};
    \node at (-1.48, 5.25) [rotate=-0, align=center, font=\scriptsize] {TAPIP3D};
    \node at (1.48, 5.25) [rotate=-0, align=center, font=\scriptsize] {CoTracker360};
    \node at (4.43, 5.25) [rotate=-0, align=center, font=\scriptsize] {TAP-Out (Ours)};
    \end{tikzpicture}
    \vspace{3mm}
    \caption{Qualitative tracking results on TAPVid360. Each column shows tracked point positions on the equirectangular view across timesteps.}
    \label{fig:comparison_results}
\end{figure}

\subsection{Motion Regime Analysis}
\label{sec:motion_regimes}

To understand how performance depends on camera motion, we partition the data into three regimes based on total camera rotation (\Cref{tab:motion_regimes}). TAP-Out achieves $\delta^{x}_{\mathrm{avg,oof}} = 0.90$ at low motion, where the scene remains similar to the observed input. Performance remains strong at 0.45 for medium motion (the dominant regime at 59.6\% of points). At high motion ($>30$\dg), accuracy drops to 0.21 as the generative model must hallucinate increasingly large unobserved regions. Future improvements in video generation fidelity should directly benefit this regime.

\subsection{Hallucination Analysis}
\label{sec:hallucination}

Since TAP-Out depends on the quality of the generated panorama, we examine when outpainting produces useful content. \Cref{fig:hallucination_robustness} compares the outpainted panorama to ground truth over time. Argus maintains global scene structure (room layout, horizon) with good fidelity, providing spatial anchoring for segmentation. For typical occlusion durations, it captures approximate object trajectories sufficient for coarse localisation. The primary failure mode is \emph{identity drift} over long occlusions, where generated content gradually deviates from the true scene. Even imperfect hallucination helps: CoTracker360 must extrapolate OOV directions without visual explanation, while TAP-Out's explicit scene generation constrains the problem more effectively. The substantial improvement in $\delta^{x}_{\mathrm{avg,oof}}$ demonstrates that approximate scene completion substantially outperforms implicit extrapolation.

\subsection{Qualitative Results}

\begin{figure}[!t]
    \centering
    \begin{tikzpicture}
    \node (img) {\includegraphics[width=0.82\textwidth, clip, trim=0 0 0 1.5cm]{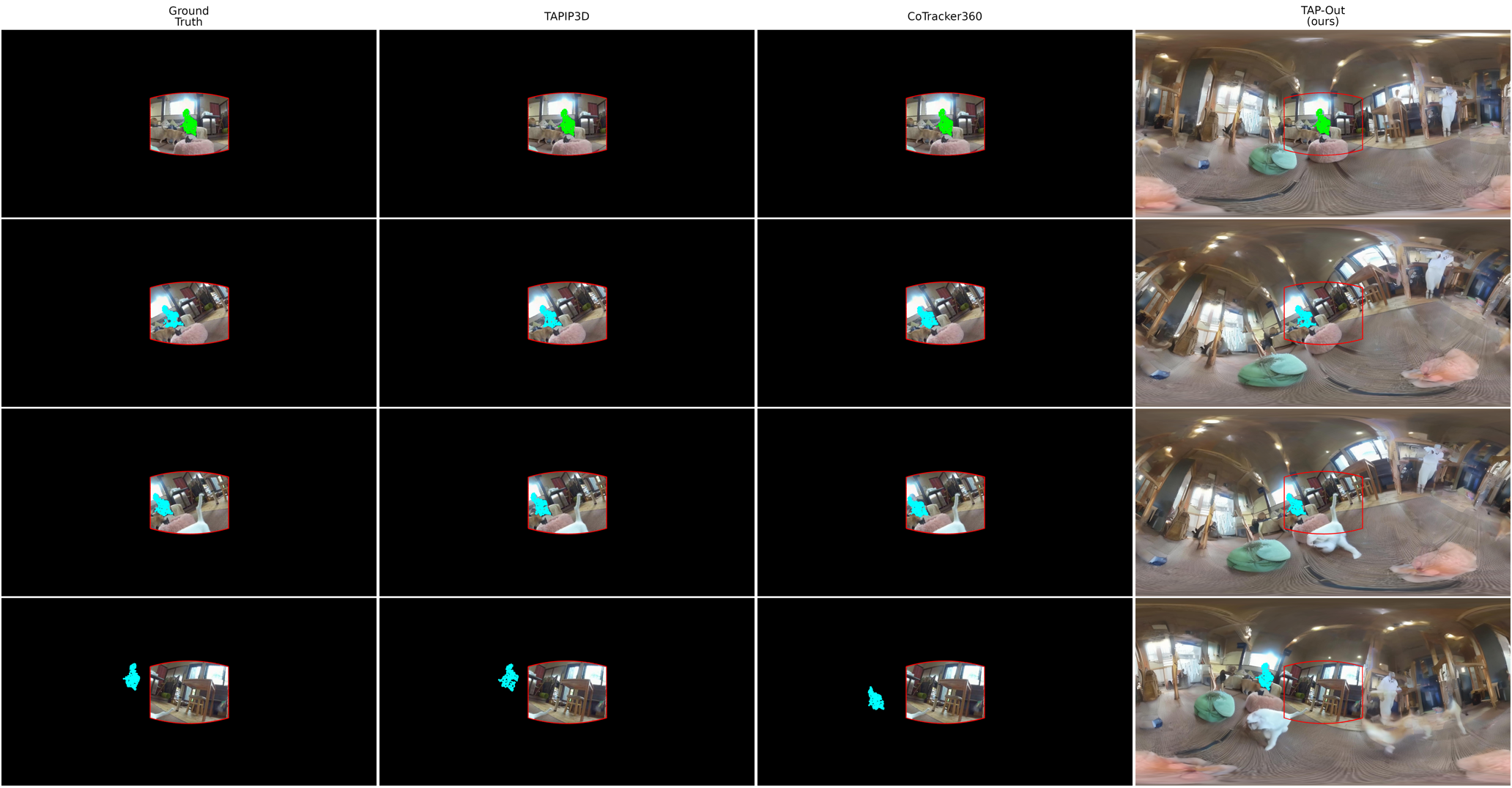}};
    \node at (-4.43, 3.12) [rotate=-0, align=center, font=\scriptsize] {Ground Truth};
    \node at (-1.48, 3.12) [rotate=-0, align=center, font=\scriptsize] {TAPIP3D};
    \node at (1.48, 3.12) [rotate=-0, align=center, font=\scriptsize] {CoTracker360};
    \node at (4.43, 3.12) [rotate=-0, align=center, font=\scriptsize] {TAP-Out (Ours)};
    \end{tikzpicture}
    \vspace{3mm}
    \caption{Further qualitative comparison on TAPVid360 with significant camera motion.}
    \label{fig:further_comparison_results}
\end{figure}

\Cref{fig:comparison_results,fig:further_comparison_results} present qualitative results on TAPVid360. Standard 2D trackers (TAPIP3D) lose track when points leave the FoV. CoTracker360 produces noisy OOV estimates that often diverge from ground truth. TAP-Out maintains coherent tracks through extended OOV periods by operating on the outpainted panoramic scene.

\subsection{Additional Tasks: Out-of-Frame Object Tracking}

\begin{figure}[!t]
    \centering
    \includegraphics[width=0.62\textwidth]{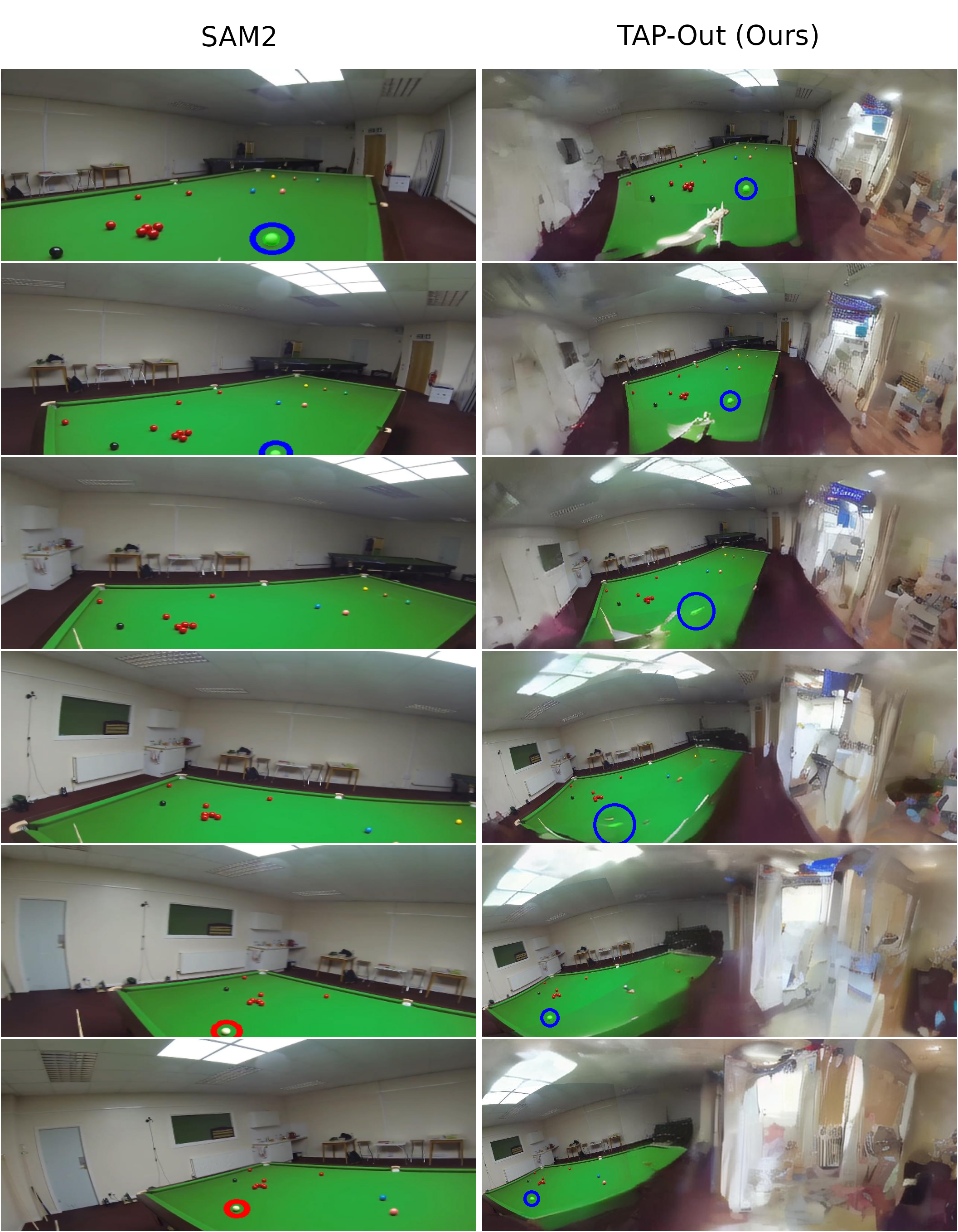}
    \vspace{3mm}
    \caption{Eliminating re-identification failures on LaSOT~\cite{fan2019lasot}. A standard tracker (SAM2, left) tracks a white ball when visible (blue circles) but fails to re-identify it upon return (red circle) after extended absence. TAP-Out (right) maintains continuous localisation in the equirectangular domain.}
    \label{fig:reid_failing}
\end{figure}

\begin{figure}[!t]
    \centering
    \resizebox{0.85\textwidth}{!}{
    \begin{tikzpicture}
    \node (img) 
    {\includegraphics[width=0.76\textwidth, clip, trim=0 0 0 1.4cm]{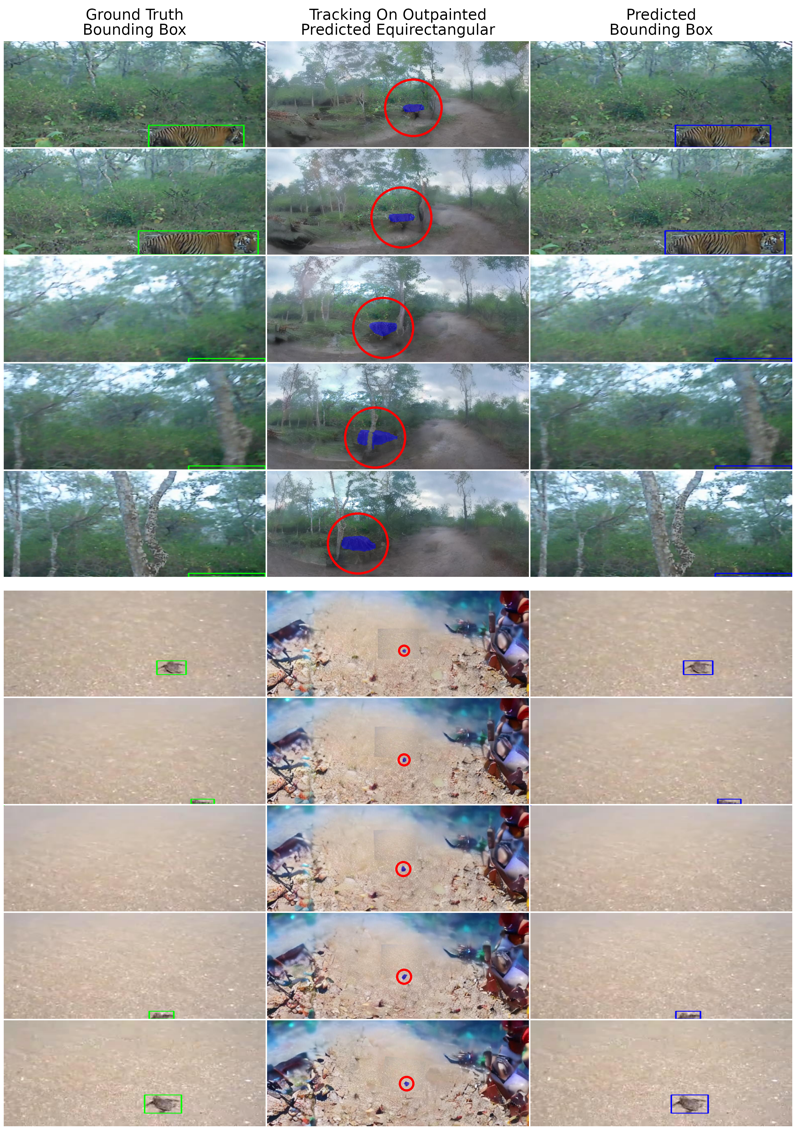}};
    \node at (-3.3, 7) [rotate=-0, align=center, font=\scriptsize] {Ground Truth \\ Bounding Box};
    \node at (-0.0, 7) [rotate=-0, align=center, font=\scriptsize] {Tracking on Outpainted \\ Equirectangular};
    \node at (3.3, 7) [rotate=-0, align=center, font=\scriptsize] {Predicted \\ Bounding Box};
    \end{tikzpicture}
    }
    \vspace{3mm}
    \caption{TAP-Out applied to out-of-frame object tracking on the LaSOT benchmark~\cite{fan2019lasot}.}
    \label{fig:lasot_outputs}
\end{figure}

We also demonstrate TAP-Out on the LaSOT benchmark~\cite{fan2019lasot}, which contains clips where objects leave and re-enter the scene, creating a challenging re-identification (Re-ID) problem. \Cref{fig:reid_failing} highlights a critical case: a white ball in a snooker clip vanishes, the camera moves, and when the ball reappears, SAM2 fails to re-identify it. TAP-Out circumvents Re-ID entirely by continuously tracking in the equirectangular domain. Further qualitative LaSOT results are provided in Figure \ref{fig:lasot_outputs} and showcase the generalised nature of the TAP-Out pipeline.

%-------------------------------------------------------------------------
\section{Discussion}
\label{sec:discussion}

\mypara{Why 360\dg{} Direction Over Explicit 3D.}
A natural alternative would be to extend 3D tracking beyond the visible frustum. However, this is fundamentally ill-posed: depth is ambiguous outside the frustum where no visual evidence constrains the estimate. Methods such as SpatialTracker~\cite{xiao2024spatialtracker} require observed depth and cannot extrapolate beyond it. The 360\dg{} formulation sidesteps this by predicting only \emph{direction} ($S^2$), which is naturally aligned with generative video models that operate in pixel space.

\mypara{Explicit vs.\ Implicit Extrapolation.}
All OOV tracking methods must extrapolate beyond observed data. CoTracker360~\cite{hudson2025tapvid} directly predicts direction vectors, implicit hallucination without visual explanation. TAP-Out makes extrapolation explicit: the generative model produces a complete visual scene, providing a human-interpretable explanation of \emph{why} each point is tracked to a particular direction. This is useful for debugging and for applications requiring transparency, and constrains the tracking problem more effectively than direct coordinate regression.

\mypara{Failure Modes.}
TAP-Out degrades in identifiable regimes: (1)~high camera motion ($>30$\dg), where the generative model must hallucinate large unobserved regions; (2)~long occlusions, where identity drift causes segmentation failures; (3)~camera estimation errors, which propagate through the pipeline; and (4)~out-of-distribution inputs, as Argus expects world-up-aligned views.

%-------------------------------------------------------------------------
\section{Conclusions}
\label{sec:conclusions}

We have presented TAP-Out, a modular pipeline that enables continuous tracking of points and objects regardless of whether they are within the camera's field of view. By decoupling in-view correspondence from OOV scene completion, and leveraging a generative video model as an explicit world-model prior, TAP-Out achieves state-of-the-art performance on the TAPVid360 benchmark with approximately $3\times$ improvement on OOV points over the best prior method.

\mypara{Limitations.}
TAP-Out is restricted to offline use (approximately 470s vs.\ 0.15s for CoTracker360). Camera estimation errors propagate through the pipeline (\Cref{tab:tracking_performance_comparison}). Argus expects world-up-aligned views, and arbitrary orientations degrade outpainting. Prolonged occlusions cause identity drift. Static points could be handled more efficiently via geometric reprojection alone.

\mypara{Future Work.}
Promising directions include distillation into a faster feed-forward model, sampling multiple outpainting hypotheses to model uncertainty, and developing online variants. The modular design ensures that improvements in any component directly benefit the overall system.

% ---- Bibliography ----
\bibliography{main}

%-------------------------------------------------------------------------
\clearpage
\appendix
\setcounter{figure}{0}
\setcounter{table}{0}
\renewcommand{\thefigure}{A\arabic{figure}}
\renewcommand{\thetable}{A\arabic{table}}

\section*{Supplementary Material}

\section{Justification of the TAPVid360-TAPOut Split}
\label{sec:split_justification}

To evaluate our method on the TAPVid360 benchmark, we utilised a subset of 8{,}887 clips from the original 10{,}000 to avoid data contamination from Argus training data. As detailed in \Cref{tab:subset_is_fine}, the results obtained on our subset (main values) exhibit negligible deviation from those of the full dataset (values in parentheses). Note that this was run on all points, not just dynamic ones. The performance metrics remain statistically consistent across all evaluated methods, and crucially, the relative ranking of the trackers is unchanged. This strongly suggests that the \textit{TAPVid360-TAPOut} split preserves the data distribution and difficulty of the original benchmark, serving as a valid and representative proxy for evaluation.

\begin{table}[!t]
  \centering
  \resizebox{\textwidth}{!}{
  % \begin{tabularpreamble}
  \begin{tabular}{lcccccc}
    \toprule
    Method & $< \delta^{x}_{\mathrm{avg}}$\textsubscript{all} $\uparrow$ & $< \delta^{x}_{\mathrm{avg}}$\textsubscript{if} $\uparrow$ & $< \delta^{x}_{\mathrm{avg}}$\textsubscript{oof} $\uparrow$ & ${AD^{x}_\mathrm{avg}}$\textsubscript{all}$\downarrow$ & ${AD^{x}_\mathrm{avg}}$\textsubscript{if} $\downarrow$ & ${AD^{x}_\mathrm{avg}}$\textsubscript{oof} $\downarrow$\\
    \midrule
TAPNext~\cite{tapnext2025} & 0.0082 (0.0082) & 0.0190 (0.0191) & 0.0004 (0.0004) & 51.98 (51.98) & 36.58 (36.59) & 62.45 (62.46) \\
TAPIR~\cite{tapir2023} & 0.0106 (0.0106) & 0.0250 (0.0251) & 0.0003 (0.0003) & 49.83 (49.81) & 33.87 (33.88) & 60.53 (60.52) \\
BootsTAPIR~\cite{bootstap2024} & 0.0125 (0.0126) & 0.0291 (0.0293) & 0.0005 (0.0005) & 48.37 (48.36) & 33.30 (33.32) & 58.38 (58.38) \\
TAPIP3D~\cite{tapip3d2025} & 0.2449 (0.2476) & 0.4685 (0.4698) & 0.0817 (0.0850) & 36.44 (36.44) & 23.08 (23.32) & 45.90 (45.80) \\
SpatialTracker~\cite{xiao2024spatialtracker} & 0.2235 (0.2239) & 0.4889 (0.4893) & 0.0300 (0.0303) & 38.90 (38.88) & 22.16 (22.16) & 50.45 (50.44) \\
CoTracker3 (offline)~\cite{cotracker2024} & 0.2444 (0.2435) & 0.5601 (0.5588) & 0.0158 (0.0158) & 37.40 (37.43) & 17.60 (17.64) & 50.94 (50.98) \\
CoTracker360~\cite{hudson2025tapvid} & 0.2389 (0.2386) & 0.4064 (0.4060) & 0.1164 (0.1160) & 8.24 (8.27) & 3.92 (3.95) & 10.93 (10.98) \\
    \bottomrule
  \end{tabular}
  }
  \vspace{3mm}
  \caption{Comparison of tracking performance on the TAPVid360-TAPOut split versus the original full TAPVid360-10k dataset (shown in parentheses).}
  \label{tab:subset_is_fine}
\end{table}

\section{Rationale for Dynamic Point Threshold}
\label{sec:dynamic_points_justification}
To categorize tracking challenges in the TAPVid360 benchmark, we define \textit{dynamic} points as those exhibiting an independent world-frame motion magnitude of at least 10\dg. This threshold is primarily motivated by the error floor of a non-dynamic tracking assumption. As shown in Table \ref{tab:tracking_performance_comparison} of the main manuscript, the \textit{Static Reprojection (Gt Camera)} baseline, which accounts for generated camera rotation but assumes objects are stationary in the world, results in a mean angular error of 8.6\dg. Consequently, point motion below 10\dg is frequently indistinguishable from the baseline noise of a static-world tracker. 

\section{2D Tracker Modularity}
\label{sec:tracker_modularity}

To validate that our framework is modular and not coupled to a specific tracking backbone, we keep the upstream camera estimation (VGGT) and outpainting modules (Argus) frozen, and swap our default \textbf{CoTracker3} backbone for \textbf{BootsTAPIR} without any retraining or fine-tuning. As shown in Table~\ref{tab:tracker_ablation}, whilst performance drops slightly when moving away from CoTracker3 (OOV accuracy decreases from $0.339$ to $0.298$), our pipeline still maintains solid out-of-frame tracking capability, confirming its plug-and-play modularity.

\begin{table}[h]
\centering
\resizebox{\columnwidth}{!}{%
\begin{tabular}{@{}llcc@{}}
\toprule
\textbf{Generative Prior} & \textbf{2D Tracker Backbone} & $\boldsymbol{\delta_{\mathrm{avg, oof}}^{x}} \uparrow$ (OOV Acc.) & $\mathbf{AD_{\mathrm{avg, oof}}^{x}} \downarrow$ (Mean Error) \\
\midrule
Argus & \textbf{CoTracker3} \textit{(Ours, Default)} & \textbf{0.339} & \textbf{5.08$^\circ$} \\
Argus & BootsTAPIR & 0.298 & 5.85$^\circ$ \\
\bottomrule
\end{tabular}%
}
\vspace{6pt}
\caption{\textbf{Plug-and-Play 2D Tracker Ablation} on a random sample of 500 videos.}
\label{tab:tracker_ablation}
\end{table}

\section{Additional Qualitative Results}

\begin{figure}[t]
    \centering
    \begin{tikzpicture}
    \node (img) {\includegraphics[width=\textwidth, clip, trim=0 0 0 1.5cm]{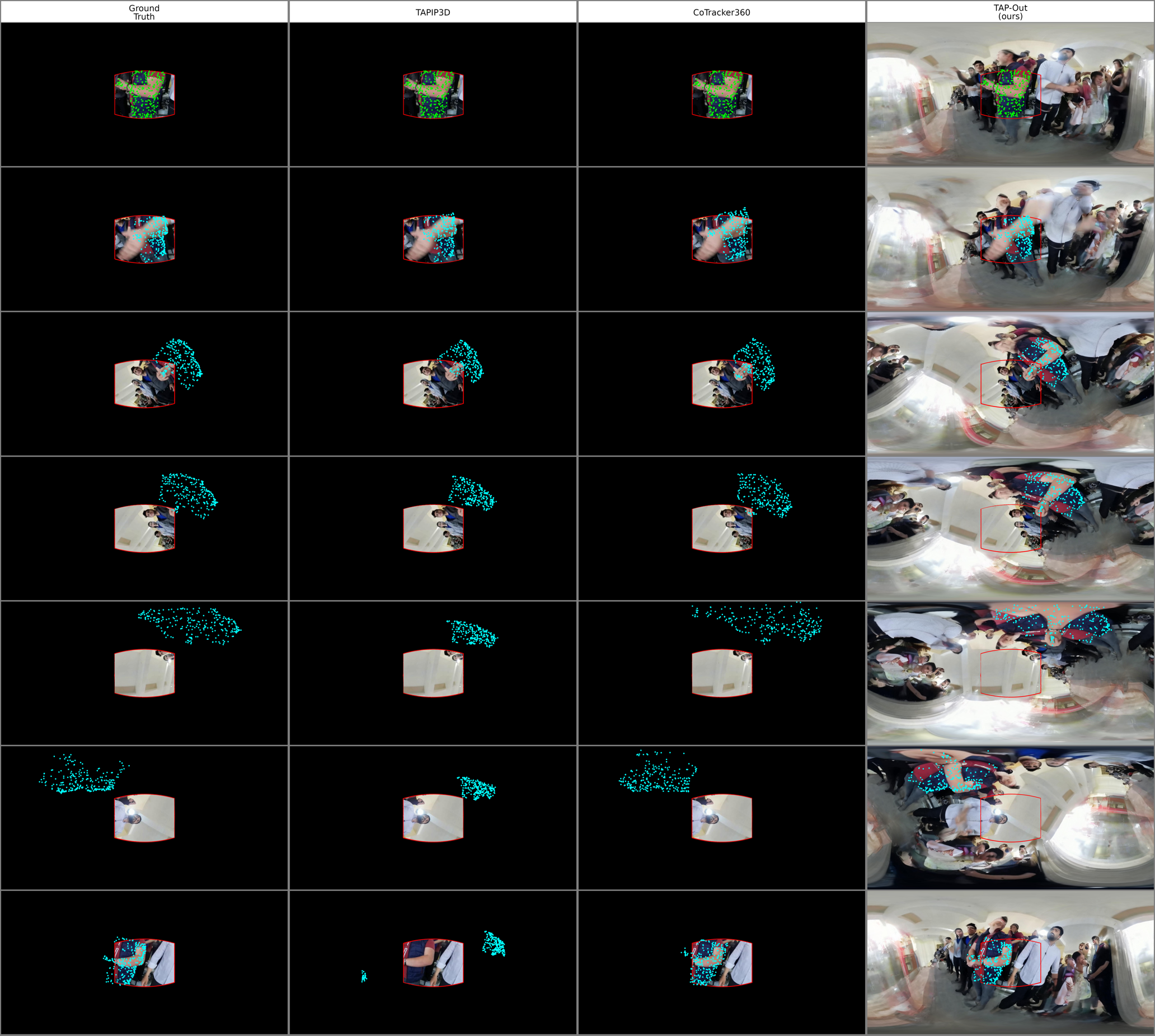}};
    \node at (-5.4, 6.4) [rotate=-0, align=center, font=\footnotesize] {Ground Truth};
    \node at (-1.8, 6.4) [rotate=-0, align=center, font=\footnotesize] {TAPIP3D};
    \node at (1.8, 6.4) [rotate=-0, align=center, font=\footnotesize] {CoTracker360};
    \node at (5.4, 6.4) [rotate=-0, align=center, font=\footnotesize] {TAP-Out (Ours)};
    \end{tikzpicture}
    \vspace{3mm}
    \caption{Additional qualitative comparison on TAPVid360 showing tracking results across methods.}
    \label{fig:more_further_comparison_results}
\end{figure}

\begin{figure}[t]
    \centering
    \begin{tikzpicture}
    \node (img) {\includegraphics[width=\textwidth, clip, trim=0 0 0 1.5cm]{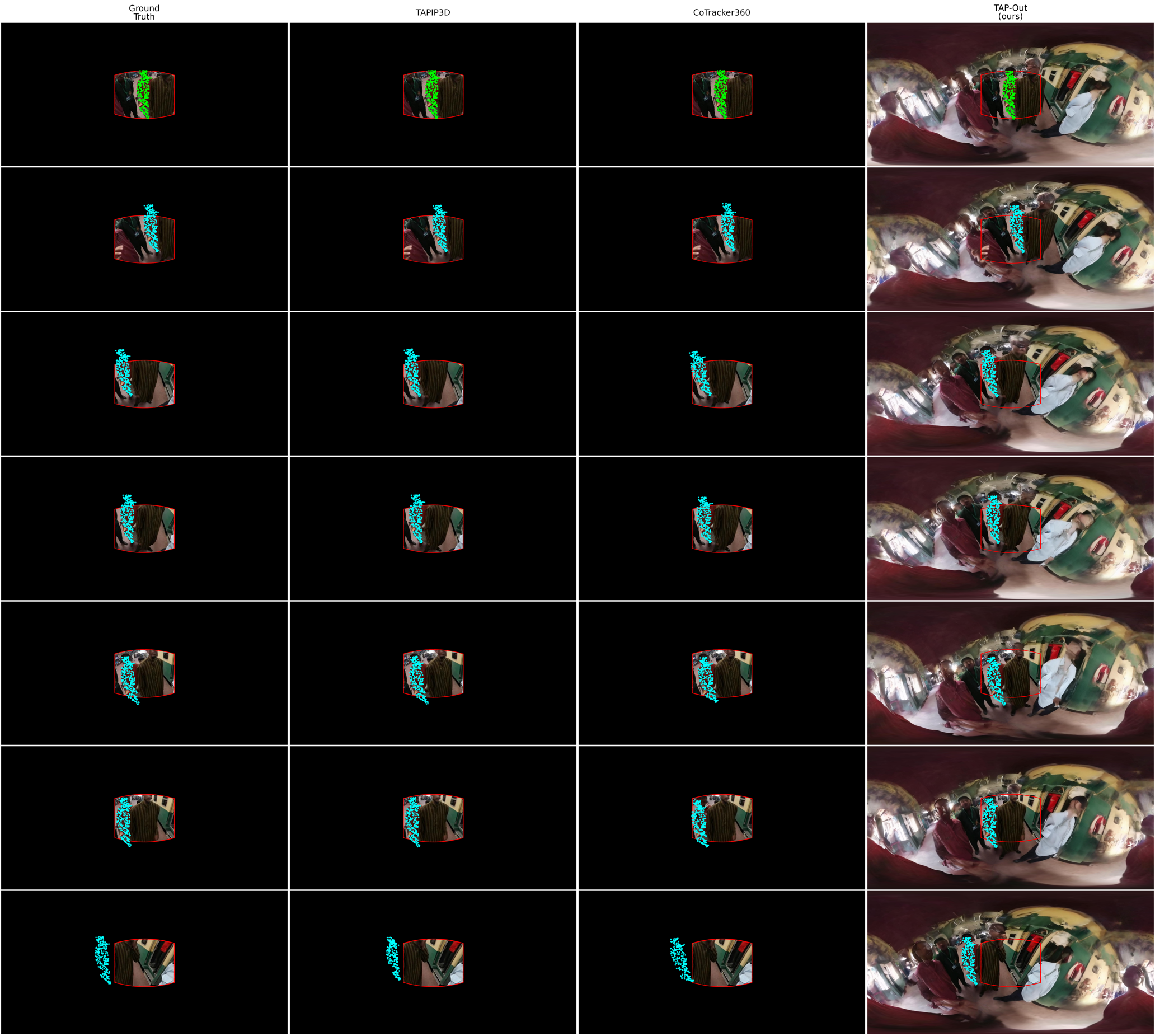}};
    \node at (-5.4, 6.4) [rotate=-0, align=center, font=\footnotesize] {Ground Truth};
    \node at (-1.8, 6.4) [rotate=-0, align=center, font=\footnotesize] {TAPIP3D};
    \node at (1.8, 6.4) [rotate=-0, align=center, font=\footnotesize] {CoTracker360};
    \node at (5.4, 6.4) [rotate=-0, align=center, font=\footnotesize] {TAP-Out (Ours)};
    \end{tikzpicture}
    \vspace{3mm}
    \caption{Further qualitative comparison on TAPVid360 showing another example sequence.}
    \label{fig:more_more_further_comparison_results}
\end{figure}

\begin{figure}[t]
    \centering
    \includegraphics[width=0.4\textwidth]{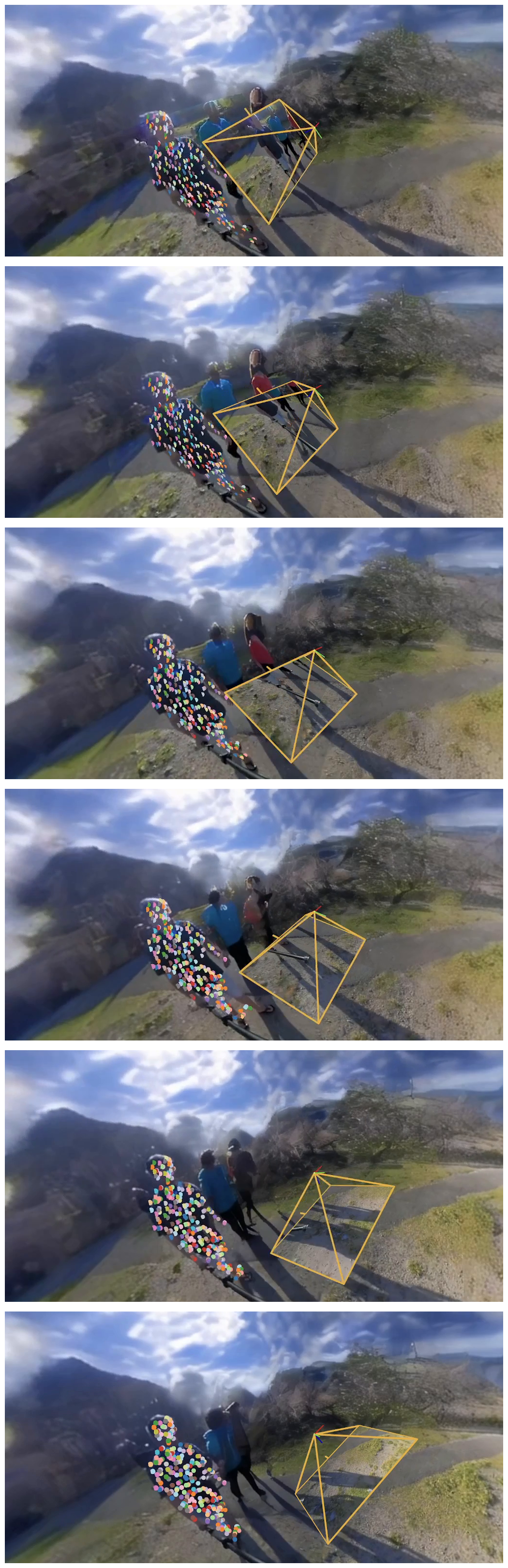}
    \vspace{3mm}
    \caption{Extended visualisation of TAP-Out results.}
    \label{fig:viser_vis}
\end{figure}

\Cref{fig:more_further_comparison_results,fig:more_more_further_comparison_results} provide additional qualitative comparisons on TAPVid360 across different scene types and motion patterns. \Cref{fig:viser_vis} shows an extended visualisation of the TAP-Out pipeline output, illustrating how the method maintains tracking consistency between the perspective and equirectangular domains.

\end{document}